\documentclass[twocolumn]{fairmeta}

\usepackage[most]{tcolorbox}

\usepackage[amssymb]{SIunits}
\usepackage{graphicx}

\usepackage{multirow}
\usepackage{amsmath,amssymb}

\usepackage{algorithm}
\usepackage{algpseudocode}
\let\labelindent\relax
\usepackage{enumitem}
\usepackage{balance}
\usepackage[misc]{ifsym}
\usepackage{booktabs}
\usepackage{lscape}
\usepackage{array}
\usepackage{balance}
\usepackage{xcolor}
\usepackage{color}
\usepackage{bm} 
\usepackage{bbm}
\usepackage{algorithm}  
\usepackage{algpseudocode}
\usepackage{multirow}
\usepackage{booktabs}
\usepackage{enumitem}
\usepackage{lscape}
\usepackage[normalem]{ulem}
\usepackage{url}
\usepackage{adjustbox}

\setlist[itemize]{noitemsep,nolistsep,leftmargin=17pt}
\setlist[enumerate]{noitemsep,nolistsep,leftmargin=17pt}
\usepackage[font={small}]{caption}
\usepackage[misc]{ifsym}
\usepackage{epigraph}
\usepackage{csquotes}
\usepackage{changepage}

\newcommand{\fref}[1]{Fig.~\ref{#1}}
\newcommand{\sref}[1]{Sec.~\ref{#1}}
\newcommand{\tref}[1]{Table~\ref{#1}}

\usepackage[colorinlistoftodos,prependcaption,textsize=tiny]{todonotes}
\title{RM-RL: Role-Model Reinforcement Learning for Precise Robot Manipulation}

\author[1,*]{Xiangyu Chen}
\author[1]{Chuhao Zhou}
\author[1]{Yuxi Liu}
\author{Jianfei Yang$^{1,\dagger}$}

\affiliation[1]{MARS Lab, Nanyang Technological University}

\abstract{
Precise robot manipulation is critical for fine-grained applications such as chemical and biological experiments, where even small errors (e.g., reagent spillage) can invalidate an entire task. Existing approaches often rely on pre-collected expert demonstrations and train policies via imitation learning (IL) or offline reinforcement learning (RL). However, obtaining high-quality demonstrations for precision tasks is difficult and time-consuming, while offline RL commonly suffers from distribution shifts and low data efficiency.
We introduce a Role-Model Reinforcement Learning (RM-RL) framework that unifies online and offline training in real-world environments. The key idea is a role-model strategy that automatically generates labels for online training data using approximately optimal actions, eliminating the need for human demonstrations. RM-RL reformulates policy learning as supervised training, reducing instability from distribution mismatch and improving efficiency. A hybrid training scheme further leverages online role-model data for offline reuse, enhancing data efficiency through repeated sampling.
Extensive experiments show that RM-RL converges faster and more stably than existing RL methods, yielding significant gains in real-world manipulation: 53\% improvement in translation accuracy and 20\% in rotation accuracy. Finally, we demonstrate the successful execution of a challenging task, precisely placing a cell plate onto a shelf, highlighting the framework’s effectiveness where prior methods fail. Project site: \url{https://ntumars.github.io/project/RMRL}
}

\correspondence{Jianfei Yang at \email{jianfei.yang@ntu.edu.sg}}

\begin{document}

\maketitle
\section{INTRODUCTION}\label{sec:introduction}
Autonomous robotic manipulation has shown promising results in real-world tasks, such as folding clothes and performing household manipulations~\cite{zitkovich2023rt, black2024pi_0, kim2024openvla}. 
Beyond these general-purpose applications, growing attention has shifted toward high-precision manipulation, particularly in delicate biological and chemical experiments~\cite{lan2025autobio, li2024chemistry3d}. In such domains, success hinges on the robot’s ability to execute actions with sub-millimeter to millimeter-level accuracy, as even minor deviations can compromise the entire experiment.

Two primary paradigms have been widely adopted for robotic policy learning: Imitation Learning (IL) and Reinforcement Learning (RL).
Imitation Learning (IL)~\cite{hussein2017imitation, ravichandar2020recent} provides an effective way to acquire policies from expert demonstrations. 
However, in high-precision tasks, even human operators struggle to deliver consistent millimeter-level accuracy through teleoperation, making the collection of demonstrations slow and costly. 
Learning from such limited data often leads to overfitting to the training distribution and poor generalization in real-world scenarios.
In contrast, Reinforcement Learning (RL) enables policies to autonomously explore and optimize through trial and error, thereby eliminating the need for demonstrations. 
Yet, most RL methods are developed and validated in simulators~\cite{geng2025roboverse, todorov2012mujoco, mittal2023orbit}, which fail to capture the subtle but critical errors in fine-grained real-world manipulation due to the persistent sim-to-real gap. 
This motivates our goal: to enable efficient training of robotic policies with high-precision manipulation skills directly in the physical world.

However, applying online RL directly in the real world is far from straightforward. In existing online RL, policies are typically updated directly from real-world reward signals. This process is highly data-inefficient, as each interaction can only be used once, and data collection in the physical world is inherently slow. Attempts to improve efficiency with replay buffers~\cite{levine2020offline, mnih2015human, lillicrap2015continuous} partially mitigate this issue but introduce a second challenge: distribution shift~\cite{rowland2019statistics, fujimoto2019off, liu2020provably}. Because stored data is generated by outdated policies, the mismatch between past experiences and the current policy often destabilizes optimization and impedes convergence. These limitations lead us to the central question of this work: \textbf{Can we design a real-world RL framework that achieves both high data efficiency and training efficiency for precise robotic manipulation?}

\begin{figure}
    \centering
    \includegraphics[width=\linewidth]{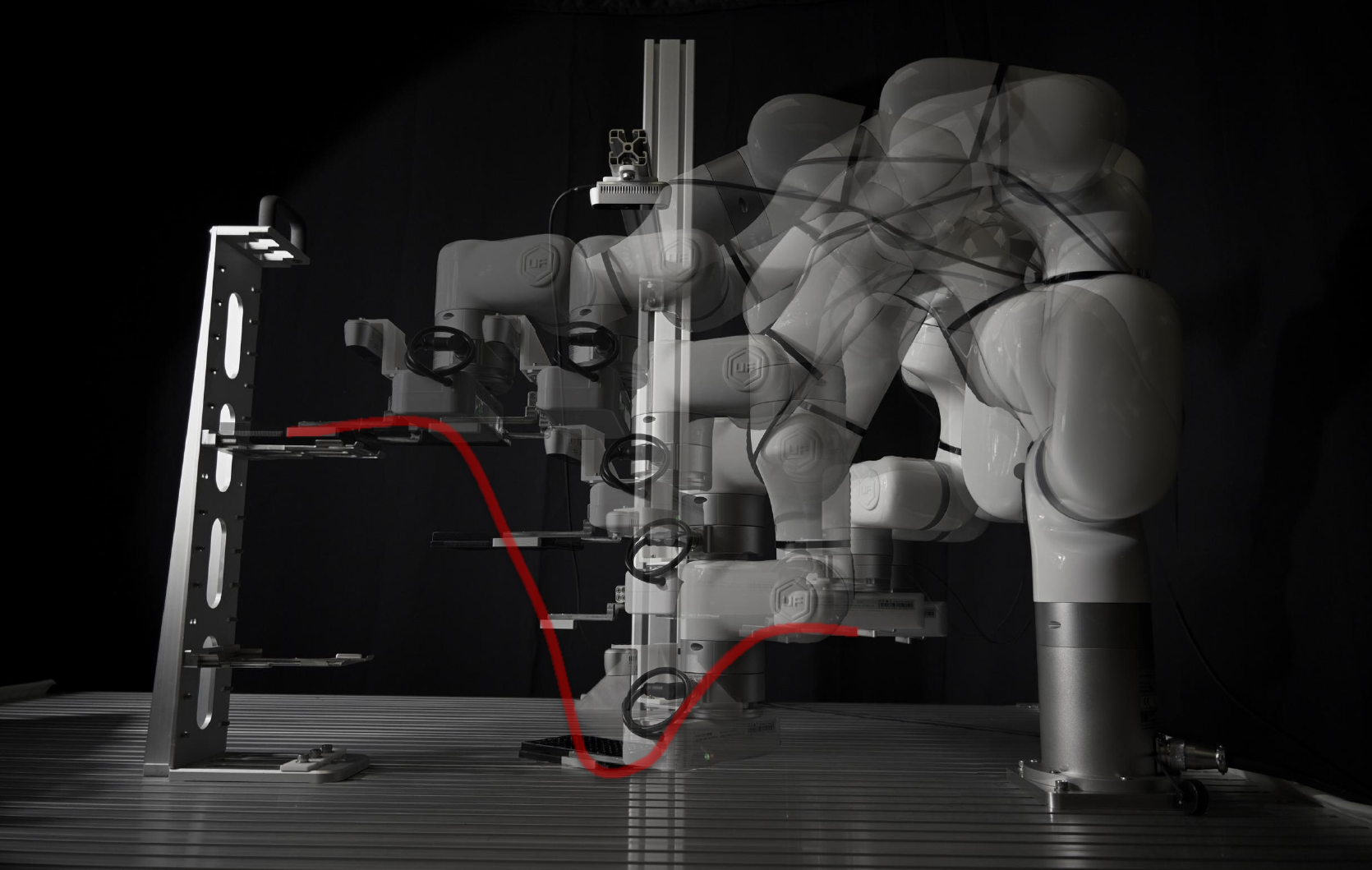}
    \caption{The Ufactory X-ARM 6 autonomously executes a pick-and-place task, transferring a cell plate to the designated shelf. The red curve illustrates the trajectory of the end-effector.}
    \label{fig:first_page}
\end{figure}

\vspace{2mm}
\begin{adjustwidth}{4pt}{4pt}
\begin{displayquote}\itshape
``When three people walk together, there must be a \underline{\textbf{role model}} whom I can learn from; I will select the good qualities and follow them." \par\vspace{-0.6em}
\rule{\linewidth}{0.4pt}
\vspace{-16pt}
\begin{flushright}
--- Confucius, ``The Analects''
\end{flushright}
\end{displayquote}
\end{adjustwidth}

To address this question, we draw inspiration from Confucius’ dictum in \textit{The Analects}, and propose the Role-Model Reinforcement Learning (RM-RL) framework. RM-RL integrates online exploration with offline supervised learning by periodically selecting a role-model action, which is the highest-reward action observed under similar initial states, and using it to guide the remaining peer actions. The peer actions are automatically labeled by reference to the role-model action, allowing the online samples to be reformulated into supervised training data. 
These labeled samples are then repeatedly reused in offline training, ensuring that valuable experiences contribute to policy improvement multiple times rather than only once. This selection-and-labeling process is performed periodically during online interaction, which enables the framework to continuously inject stable supervised signals into the training loop. 
As a result, RM-RL substantially enhances data efficiency and mitigates distribution mismatch~\cite{kumar2022should, sarker2021machine}, ultimately achieving both stable optimization and robust policy learning for high-precision robotic manipulation in the real world.

To validate the effectiveness of the RM-RL framework in real-world applications, we consider a precise manipulation task: placing a cell plate into a designated slot with millimeter-level accuracy, such as a cell plate shelf, as shown in~\fref{fig:first_page}.
At each iteration, the policy network predicts actions based on the real-world environmental observation, which are then transformed to control commands for the robotic arm to execute. 
The reward is computed as the deviation between the executed and target poses and used to update the policy network via policy gradient.
The role-model samples are simultaneously identified to label online training data, enabling their reuse during offline training. 
The offline training operates in a supervised manner, iteratively fine-tuning the policy network to improve performance.
The contributions of this paper are summarized as follows:

\begin{itemize}
    \item We propose a role-model mechanism that, by sampling under similar initial conditions, identifies approximately optimal actions to label real-world samples collected during online RL training. This labeling process transforms offline learning into a supervised paradigm, thereby enhancing the efficiency of real-world RL training.
    \item We propose a combined online–offline RL framework in which data are collected during online training and subsequently labeled through the role-model mechanism. Within this framework, each sample obtained from online training is reused multiple times in the offline training stage, thereby enhancing data efficiency.
    \item The experiment results demonstrate that the proposed role-model strategy and recipe can improve the sampling data efficiency, get faster convergence, and achieve a better success rate in the real-world, precise tasks.
\end{itemize}


\section{Related Works}\label{sec:related works}
\subsection{Reinforcement Learning}
Reinforcement learning (RL) aims to train a policy to make decisions based on the reward generated from the environment, typically modeled as a Markov decision process (MDP)~\cite{puterman2014markov}. Over the past decade, RL has demonstrated its power in combination with deep neural networks in multi-agent systems~\cite {busoniu2008comprehensive} and robotic areas, such as autonomous driving and robot locomotion tasks. However, RL still faces challenges when using real-world data for training~\cite{dulac2019challenges}.

Traditional RL methods can be broadly categorized as online RL, where the policy model is updated simultaneously with the environment's reward during the training process. 
Due to the emergence of high-performance simulators, like Isaaclab~\cite{mittal2023orbit}, Mujoco~\cite{todorov2012mujoco}, online RL has demonstrated strong power in robotic locomotion, with Q value-based approaches (e.g., Q-learning and DQN~\cite{mnih2015human}), policy gradient methods~\cite{williams1992simple}, and actor-critic frameworks (e.g., A3C~\cite{mnih2016asynchronous}, SAC~\cite{degris2012model}). Despite their success, online approaches often suffer from high sample complexity and safety concerns, making their deployment in real-world systems such as robotics and autonomous driving particularly challenging.

To address the limitations of costly and risky online interaction, offline RL (also referred to as batch RL) has gained increasing attention. In offline RL, policies are trained entirely from previously collected datasets without additional environment access, allowing the reuse of experiences from demonstrations, simulators, or operational logs. Recent advances, including CQL~\cite{kumar2020conservative}, and IQL~\cite{kostrikov2021offline}, have made notable progress toward improving stability and mitigating the risk of extrapolation errors, showing promising results in safety-critical domains such as healthcare, recommendation, and robot manipulation. 
Specifically, Zhou et al.~\cite{zhou2022real} leverage an extensive dataset to improve the generalization capabilities.
Nevertheless, offline RL still faces significant challenges in practice. Widely used offline RL datasets~\cite{james2020rlbench, fu2020d4rl, rafailov2024d5rl} are almost collected in simulation environments, where there exist significant gaps between the real-world settings. 
Collecting large-scale and high-quality real-world robotic datasets \cite{walke2023bridgedata} remains difficult and resource-intensive, while the issue of distributional shift persists, as learned policies may generate out-of-distribution actions unsupported by the dataset. These limitations continue to hinder the scalability and reliability of offline RL in real-world applications. 

\subsection{ Reinforcement Learning for Real-world Robotic Tasks}

Reinforcement Learning (RL) has demonstrated remarkable capabilities in robotic domains such as locomotion, manipulation, and autonomous navigation. Quadruped and humanoid robots have successfully employed deep RL to achieve agile locomotion directly on hardware~\cite{hwangbo2019learning, radosavovic2024real}, while mobile robots have leveraged RL for navigation in unstructured environments~\cite{wijayathunga2023challenges, liu2020robot}. In manipulation, RL has enabled large-scale robotic grasping~\cite{xu2023unidexgrasp, wan2023unidexgrasp++, geng2023rlafford}, underscoring its potential for practical deployment.
Despite these successes, deploying RL in the real world remains challenging~\cite{dulac2019challenges} due to issues of sample inefficiency, safety, and robustness to domain shifts. Collecting large amounts of on-robot data is expensive and risky, and simulators often fail to capture fine-grained physical properties, resulting in significant sim-to-real gaps~\cite{zhao2020sim, muratore2022robot}. To address these challenges, researchers have developed approaches such as sim-to-real transfer with domain randomization~\cite{tobin2017domain}, model-based RL with latent world models for efficient data usage~\cite{hafner2019dream}, and offline RL leveraging prior robot datasets for safe policy training~\cite{kostrikov2021offline}. Moreover, hybrid frameworks that combine demonstrations with online fine-tuning~\cite{nair2020awac} or integrate classical controllers with RL~\cite{johannink2019residual} further reduce the interaction burden and improve training stability.

\section{Methodology}\label{sec:method}

\vspace{4pt}
\begin{figure*}
    \centering
    \includegraphics[width=0.95\linewidth]{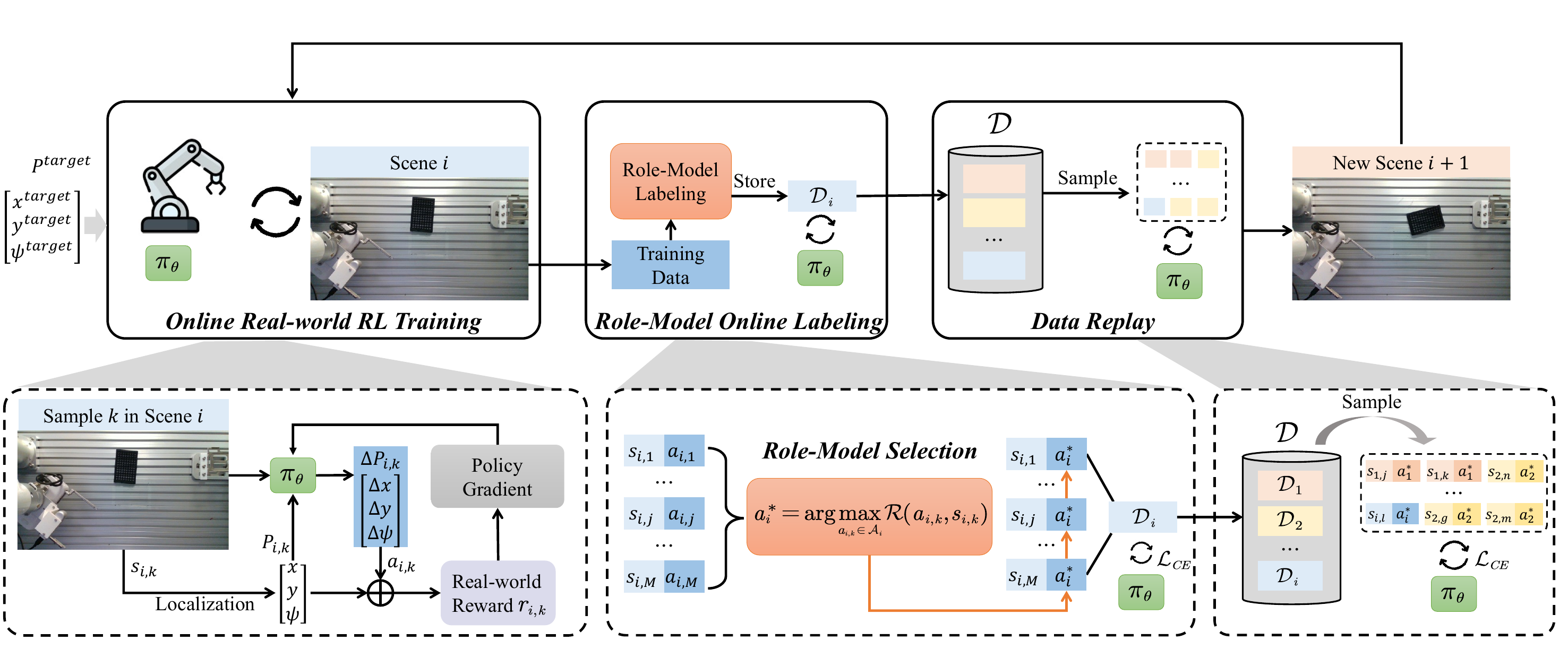}
    \caption{The framework of the proposed Role-Model Reinforcement Learning (RM-RL). The framework mainly includes three parts: the Online Real-world RL training, the Role-Model Online Labeling, and the Data Replay. In the online real-world RL part, actions are sampled to update the policy network using a policy gradient under similar scenes and initial states. In the role-model online labeling part, the approximately optimal action $a_i^*$ is selected as a label for the rest of the states. The labeled data are collected as the sub-dataset $\mathcal{D}_i$ to build a dataset $\mathcal{D}$ for further training. In the data replay part, actions and states are sampled from the $\mathcal{D}$ to update the policy network. }
    \label{fig:framework}
\end{figure*}

\subsection{Problem Definition}
As we mentioned in~\sref{sec:introduction}, our work addresses the robotic manipulating task requiring millimeter-level accuracy. 
The task involves using the robotic arm’s bio-gripper to pick and place a cell plate into a designated position of comparable size.
We formalize the task as a standard one-step Markov Decision Process (MDP)~\cite{puterman2014markov}, $\mathcal{M} = \big(\mathcal{S}, \mathcal{A}, p, \mathcal{R}, \rho_0, \gamma\big)$. Specifically, the state $\mathcal{S}$ includes the environmental images $I$ and the corresponding estimated picking poses $P^e$ of the robotic arm. The action $\mathcal{A}$ is the set of all candidate pose adjustments $\Delta P$ to $P^e$. 
For clarity, we denote each pose adjustment $\Delta P$ as an action $a\in \mathcal{A}$. 
The reward function $\mathcal{R}$ provides the real-world reward as the deviation between the final and target poses.
Since the selected task is to pick and place the cell plate on a planar surface, we only consider the pose components $x$, $y$, and $\psi$, denoted as $P = [x, y, \psi]$. Here, $x$ and $y$ represent translations along the $x$- and $y$-axes, and $\psi$ denotes the rotation about the $z$-axis. 
Thus, the action can be simplified to $a=[\Delta x, \Delta y, \Delta \psi]$. 


\subsection{Overall Framework}
\fref{fig:framework} illustrates the proposed framework of the proposed Role-Model Reinforcement Learning (RM-RL). 
The framework adopts the hybrid online-offline training paradigm and consists of three main components: Online Real-World RL Training, Role-Model Online Labeling, and Data Replay. 
Before training begins, the cell plate is placed at the target area and its pose is recorded as the target pose $P^{target}$, which remains fixed throughout the process. 
During Online Real-World RL Training, actions are sampled under similar initial states in scene $i$ and used to update the policy network $\pi_\theta$ via policy gradient.
At step $t$, the global camera captures a color image $I_t$, from which the estimated cell plate pose $P^e_{i,t}$ is obtained.
The pair $(I_{i,t}, P^e_{i,t})$ is then fed into the policy network to predict an adjustment pose $\Delta P_{i,t}$, which corresponds to the action $a_{i,t}$ defined above. 
The robotic arm executes the final picking pose, which is defined as:
\begin{equation}
    P^{pick}_{i, t} = P^e_{i, t} + a_{i, t}.
\end{equation}
Once the plate is successfully grasped, the robotic arm follows a pre-planned trajectory to place it at the center of the target area.
The resulting pose $P^{final}_{i,t}$ is recorded, and the reward is computed by the reward function $\mathcal{R}$, based on the deviation between $P^{final}_t$ and $P^{target}$. 
This reward is used to update the policy network with the policy gradient algorithm~\cite{williams1992simple}.
In the Role-Model Online Labeling, we leverage the proposed role-model strategy to online label the training data from scene $i$ as $\mathcal{D}_i$, incrementally building a well-labeled dataset $\mathcal{D}$ for further offline RL training. 
The offline training processes are executed multiple times in the Role-Model Online Labeling and Data Replay.
The details of the core contributions, role-model online labeling and online-offline training recipe, are presented in the following section. 

\subsection{Role-Model Reinforcement Learning}
This part illustrates the Role-Model Reinforcement Learning (RM-RL) in detail, including the role-model online labeling and the hybrid online-offline RL training. 
Role-model labeling enables self-annotation to improve training efficiency, while hybrid online–offline RL training emphasizes reusing scarce training data to enhance data efficiency.

\subsubsection{Role-Model Online Labeling}
The role-model online labeling selects the role-model action to label the corresponding samples within each scene in the online training stage.
During the step-by-step online training, we group the steps with similar initial states $\mathcal{S}_{i}$, their rewards $\mathcal{R}_{i}$, and actions $\mathcal{A}_{i}$ to formulate the scene $i$.
The role-model strategy is to select the action with the best reward as the role-model action $a^*_{i}$ from $\mathcal{A}_i$, which can be formulated as follows:
\begin{equation}
    a^{*}_i = \arg\max_{a_{i,k} \in \mathcal{A}_i} \mathcal{R}(a_{i,k}, s_{i,k}), 
\end{equation}
where $a_{i,k}$ and $s_{i,k}$ are the action and state in scene $i$ at step $k$.
Then, we can use the role-model action $a^*_i$ as the approximately optimal action to label the remaining states in scene $i$. 
Since the pose adjustment prediction is formulated as a discrete probability distribution, the variables $\Delta x$, $\Delta y$, and $\Delta \psi$ are restricted to a fixed set of candidate values. 
Consequently, the selection of the pose adjustments can be transformed into a multi-class classification problem, 
where predictions of $\Delta x, \Delta y$, and $\Delta \psi$ can be mapped into corresponding discrete classes, respectively.
The indices of the $\Delta x, \Delta y$ and $\Delta \psi$ of the role-model action $a_i^*$ are extracted as $\mathcal{I}_i = [ind_x, ind_y, ind_\psi]$.
Then, the states are collected in scene $i$ and labeled with $\mathcal{I}_i$ to incrementally build the dataset $\mathcal{D}$. The dataset collection process can be represented as:
\begin{align}
    \mathcal{D}_i &= \{(s_{i,k}), \mathcal{I}_i\}_{k=1}^N, \\
    \mathcal{D}   &= \{\mathcal{D}_1, \ldots, \mathcal{D}_i\},
\end{align}
where $\mathcal{D}_i$ and $\mathcal{I}_i$ represent the dataset and the index label for the $i$-th scene with similar initial states, $N$ represents the total number of the state and $s^k_i$ is one state in $i$-th scene.  
The well-labeled dataset $\mathcal{D}$ can be regarded as demonstrations for executing supervised learning to update the policy network. 

\subsubsection{Hybrid Online-Offline RL Training}
To enhance training efficiency, we adopt a hybrid online–offline framework that facilitates the effective reuse of sampled data. 
In online training, the robotic arm interacts with the real-world environment in real-time to explore and update the policy network $\pi_\theta$, while the offline training further fine-tunes the policy using role-model–labeled data.
We elaborate the details about online training, offline training, network architecture, and reward design in the following.

\textbf{Online Training}:
The online RL adopts the policy gradient~\cite{sutton1999policy} fashion to train the policy network, since other training methods, like Actor-Critic~\cite{grondman2012survey, haarnoja2018soft}, Proximal Policy Optimization~\cite{schulman2017proximal}, require extensive action samples to achieve convergence, which is time-consuming in the real world. 
Online RL aims to maximize the following objective:
\begin{equation}
    J(\theta) = \mathbb{E}_{a \sim \pi_\theta} \Big[ R(a) \Big],
\end{equation}
where $a$ is the sampled action given policy $\pi_{\theta}$ and $R(a)$ is the one-step reward of action $a$.
In policy gradient, we instead minimize the surrogate loss function for training:  
\begin{equation}
    \mathcal{L}(\theta) = - \mathbb{E}_{s,a} \Big[ \log \pi_\theta(a \mid s) \, R(a) \Big].
\end{equation}

\textbf{Offline Training}:
To enhance the data efficiency and generalization capability of the policy, we separately perform offline RL training in the Role-Model Online Labeling and Data Replay. 
Both offline RL training adopt the same loss function, while the distinction lies in the datasets. The first training is conducted on $\mathcal{D}_i$, and the second on $\mathcal{D}$ periodically. 
After collecting $\mathcal{D}_i$, the first offline training is applied, allowing the policy to learn the approximately optimal solution in the $i$-th scene.
We then perform the second offline training at fixed step intervals, ensuring that the policy periodically replays the information from the whole dataset $\mathcal{D}$. The training data are randomly sampled from the $\mathcal{D}$ during the training process.


Specifically, given a state $s$ from scene $j$ and its indices $\mathcal{I}_j$ in $\mathcal{D}$, the corresponding optimal action is $a^* = \mathcal{A}_j(ind_x, ind_y, ind_\psi)$. Consequently, the action distribution specializes to a deterministic form for $x$, $y$, and $\psi$ dimensions. For the $i$-th action dimension, the probability of the $a^*[i]$ is 1, while the probabilities of other candidate actions are set to 0. This yields a one-hot distribution in the action space for each dimension, expressed as:
\begin{equation}
\pi^*(a[i] \mid s) =
\begin{cases}
1, & \text{if } a[i] = a^*[i], \\
0, & \text{if } a[i] \neq a^*[i].
\end{cases}
\quad i=0,1,2.
\end{equation}
Then, the cross-entropy loss can be calculated as:
\begin{equation}
    \mathcal{L}_\theta = -\sum_{i=0}^2  \log \pi_\theta(a^*[i]\mid s).
\end{equation}
Additionally, this offline training scheme can be applied during a pre-training stage, where the policy network is updated using data collected from prior real-world experiments. The resulting pretrained policy offers a favorable initialization for subsequent reinforcement learning, leading to faster convergence and improved training stability.

\begin{figure}
    \centering
    \includegraphics[width=0.95\linewidth]{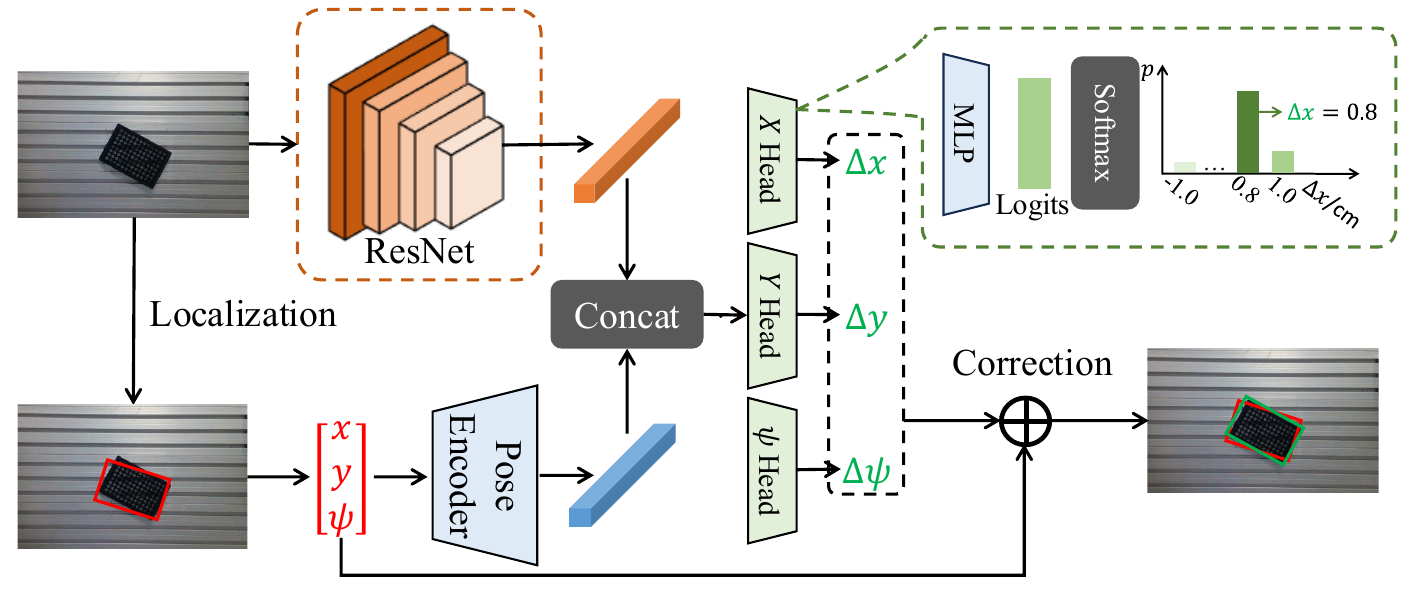}
    \caption{The Policy network. The network takes the global image and the estimated picking pose as inputs, and outputs probability distributions over $\Delta x$, $\Delta y$, and $\Delta \psi$. The final pose adjustments are sampled from these distributions to refine the picking pose.}
    \label{fig:network}
\end{figure}

\textbf{Network Design}:
The architecture of our policy network $\pi_\theta$ is shown in \fref{fig:network}. There are two inputs of the policy network, the color image $I$ and the estimated pose $P^e$. 
The output of the policy network is the adjustment pose $\Delta P$, referred to as the action $a\in\mathcal{A}$ introduced earlier, adjusting $P^e$ for precise picking. ResNet~\cite{he2016deep} is leveraged to extract the image feature, and an MLP is used to encode the input pose. 
Then, the image and pose features are concatenated and fed into three separate output heads, which respectively predict the translation adjustments along the $x$- and $y$-axes, and the rotation adjustment around the $z$-axis.
All the heads adopt the same MLP architecture to predict logits, which are transformed into probability distributions using the softmax function. 
The adjustments $\Delta x$, $\Delta y$, and $\Delta \psi$, sampled from the predicted distributions, are represented collectively as the action $a=[\Delta x, \Delta y, \Delta \psi]$.
Finally, the final picking pose is calculated as $P^{pick} = P^e + a$, which is executed by the robotic arm to pick the cell plate precisely.

\textbf{Reward Design}:
The reward is designed to evaluate the similarity between the target pose $P^{target}$ and the final pose $P^{final}$ after executing the actions.
The similarity measure separately accounts for translation and rotation errors.
For translation, the error $e_{trans}$ is defined as the Euclidean distance between the translation in the final pose $t^{final}$ and the target pose $t^{target}$, which can be formulated as follows:
\begin{equation}
    e_{trans} = \|t^{final} - t^{target}\|_2.
\end{equation}
For rotation, since the task only requires planar alignment, we only consider the yaw angle in the final and target pose for error calculation.
The cosine value of the angle difference is used to measure the orientation similarity, and the corresponding rotation error can be formulated as:
\begin{equation}
    e_{rot} = 1 - \cos(\psi^{final} - \psi^{target}) . 
\end{equation}
The overall RL reward combines the translation and rotation error, which can be calculated as:
\begin{equation}
    r = \exp(-(e_{trans}+e_{rot})).
\end{equation}
The exponential operation is applied to normalize the final reward within the range $[0,1]$, where the reward increases as the combined translation and rotation errors decrease.
In the real-world experiment, we first put the cell plate in the target position and use the overhead camera to localize it, getting the $t^{target}$ and $\psi^{target}$. 
After the robotic arm finishes executing actions, we re-localize the cell plate to get the $t^{final}$ and $\psi^{final}$ and calculate the reward. 

In summary, the Role-Model Reinforcement Learning (RM-RL) framework integrates online and offline training through a role-model strategy that provides labeled data for offline training. Unlike traditional RL with replay buffers~\cite{mnih2015human}, which suffers from distribution mismatch~\cite{levine2020offline, fujimoto2019off}, 
RM-RL constrains the policy network $\pi_\theta$ to move forward to the approximately optimal action among similar initial states by supervised learning, significantly improving sampling efficiency, adaptability, and stability.




\subsection{Cell Plate Localization}

To get the estimated pose $P^e$ of the cell plate from the real-world observation, a general pipeline consisting of object detection, segmentation, and pose estimation is leveraged. 
For precisely localizing the cell plate, we use a global camera instead of some physical labels, providing both the environmental color image $I_{RGB}$ and depth image $I_{depth}$. 
The Grounding Dino~\cite{liu2024grounding}, a pretrained open-vocabulary object detection method, is then utilized to localize the bounding box of the plate $(x_{min}, y_{min}, x_{max}, y_{max})$ via the prompt ``black cell plate''.
Furthermore, we adopt Segment Anything Model (SAM)~\cite{kirillov2023segany}, a foundation model of object segmentation, to predict the mask area $\mathcal{M}$ from the bounding box of the target cell plate.
The central pixel coordinate is denoted as $(u, v)$, where $u = \frac{x_{min} + x_{max}}{2}$ and $v = \frac{y_{min}+y_{max}}{2}$. Then, the depth value is the average depth within the cell plate mask, which can be calculated as:
\begin{equation}
    Z = mean(I_{depth}\odot\mathcal{M}),
\end{equation}
where $\odot$ represents the Hadamard product. Then, the 3D position $\mathbf{p}_{camera}= [X, Y, Z]$ of the cell plate in the camera coordinate system can be formulated as:
\begin{equation}
    X = \frac{(u - c_x)\cdot Z}{f_x}, Y = \frac{(u - c_y)\cdot Z}{f_y}, Z = Z,
\end{equation}
where $(f_x, f_y)$ are the focal lengths in pixel units and $(c_x, c_y)$ is the principal point. After recovering the $\mathbf{p}_{camera}$ in the camera coordinate system, its position in the world coordinate system is obtained using the camera extrinsic parameters. With rotation matrix $R$ and translation vector $t$, its position $\mathbf{p}_{world}$ in the world coordinate system is:
\begin{equation}
    \mathbf{p}_{world} = R\cdot \mathbf{p}_{camera} + t.
\end{equation}
Then, the yaw angle $\psi$ can be estimated from $\mathcal{M}$ by applying Principal Component Analysis (PCA)~\cite{abdi2010principal}, where the first principal component indicates the major directions. Finally, we use the combination of $\mathbf{p}_{world}$ and $\psi$ as the pick pose of the end-effector in the robotic arm to finish the picking task.

\begin{figure}[t]
    \vspace{5pt}
    \centering
    \includegraphics[width=\linewidth]{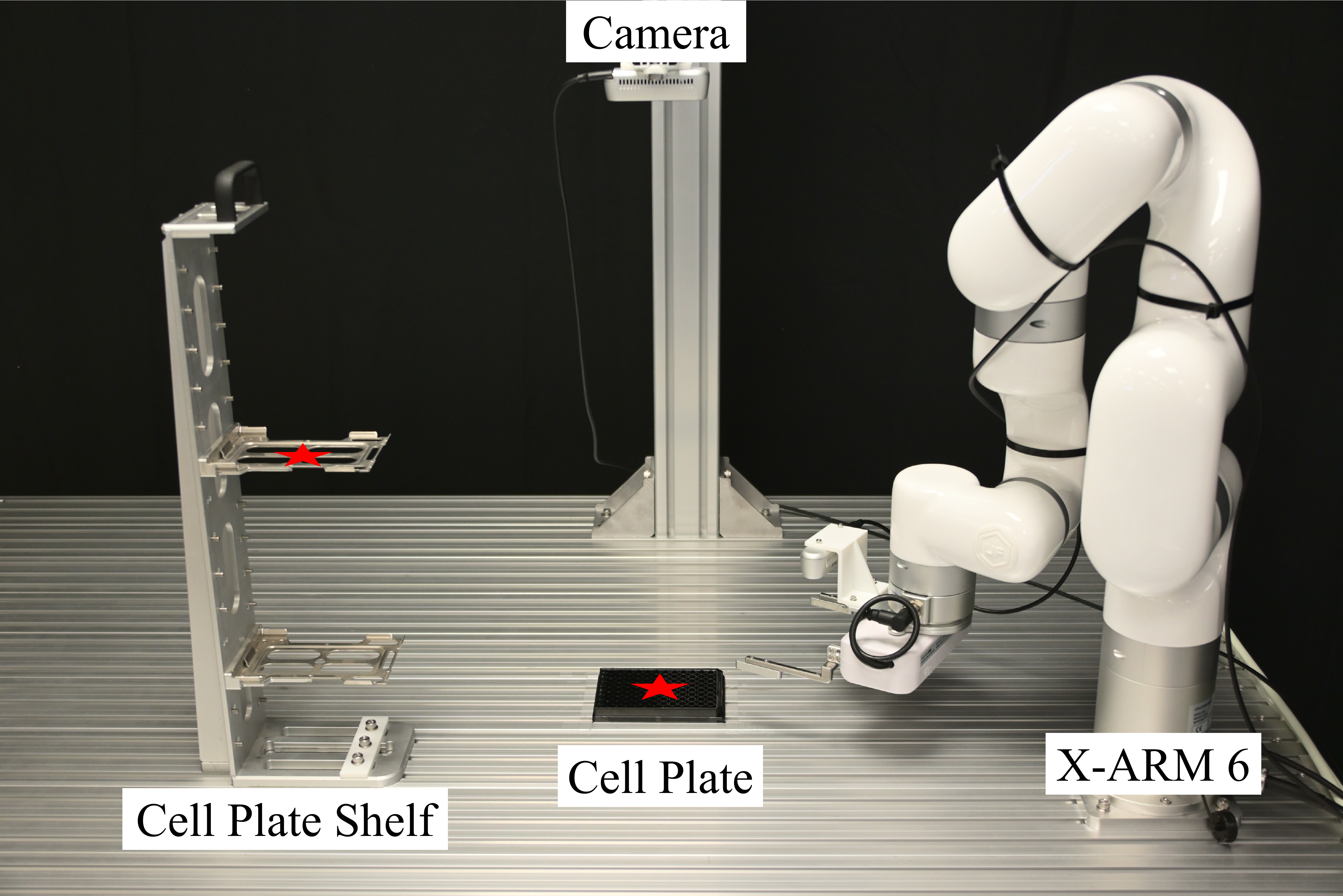}
    \caption{The real-world experimental setting. The setting includes robotic arm (X-ARM 6), cell plate, cell plate shelf, and overhead camera (Intel RealSense D435). The red stars represent the target positions for the precise picking and placing tasks.}
    \label{fig:exp: experimental hardware}
\end{figure}

\section{Experiments}\label{sec:experiment}
The experiment consists of two main parts. 
First, we compare the training performance of our framework with other RL methods, reflecting the effectiveness of the proposed role-model mechanism in improving data efficiency and training efficiency. 
Second, we evaluate the performance of the selected methods in two real-world tasks, requiring precise picking and placing. These real-world experiments further demonstrate that the proposed method can be effectively applied to practical robotic applications.

\subsection{Hardware}
For the hardware, as illustrated in \fref{fig:exp: experimental hardware}, we use a 6-DOF Ufactory X-ARM 6 with a bio gripper as our platform to finish the designed experiments. 
RealSense D435 is mounted overhead to provide depth and RGB visual perception, enabling the system to detect and localize the cell plate in the workspace. 
Two target positions, marked as red stars, are defined in the setup: one is located on the table surface (where the cell plate is placed), and the other is on the shelf slot. 
This configuration enables evaluation of reinforcement learning–based manipulation policies under real-world conditions, requiring accurate transfers between different spatial levels. For the calibration of the overhead RealSense camera, we first calibrate the camera on the robotic wrist through chessboard calibration method~\cite{zhang2002flexible} to get the transformation matrix $T^{wrist}_{world}$. Then, the wrist camera and the overhead camera localize the chessboard simultaneously, to get the transformation matrix $T_{wrist}^{camera}$.
Finally, we get the transformation matrix $T^{camera}_{world} = T_{wrist}^{camera}\cdot T^{wrist}_{world}$.

\subsection{Baseline}
To verify the effectiveness of the proposed method, we compare the performance of the standard RL~\cite{williams1992simple}, standard RL with replay buffer (RL + Replay Buffer)~\cite{mnih2015human}, the proposed Role-Model Reinforcement Learning (RM-RL), and pretrained RM-RL. 
To ensure fair comparisons, all RL methods use the same policy network and are trained with the same policy gradient-based method and training configurations.
Since the selected task is relatively simple, we adopt policy gradient for optimization instead of more advanced algorithms such as PPO or Actor-Critic. 
It is worth noting that our proposed role-model strategy is orthogonal to these algorithms and can be readily integrated into them in future work.
For the pretrained RL + MR, we collect about 200 samples, which are labeled through the proposed role-model strategy, and the policy model is trained before the beginning of real-world RL training.
To ensure consistency, all policy networks were trained and evaluated on the same PC, equipped with an NVIDIA RTX 4060 GPU (6 GB).

\subsection{Training Performance}
\begin{figure}[t]
    \vspace{5pt}
    \centering
    \includegraphics[width=0.9\linewidth]{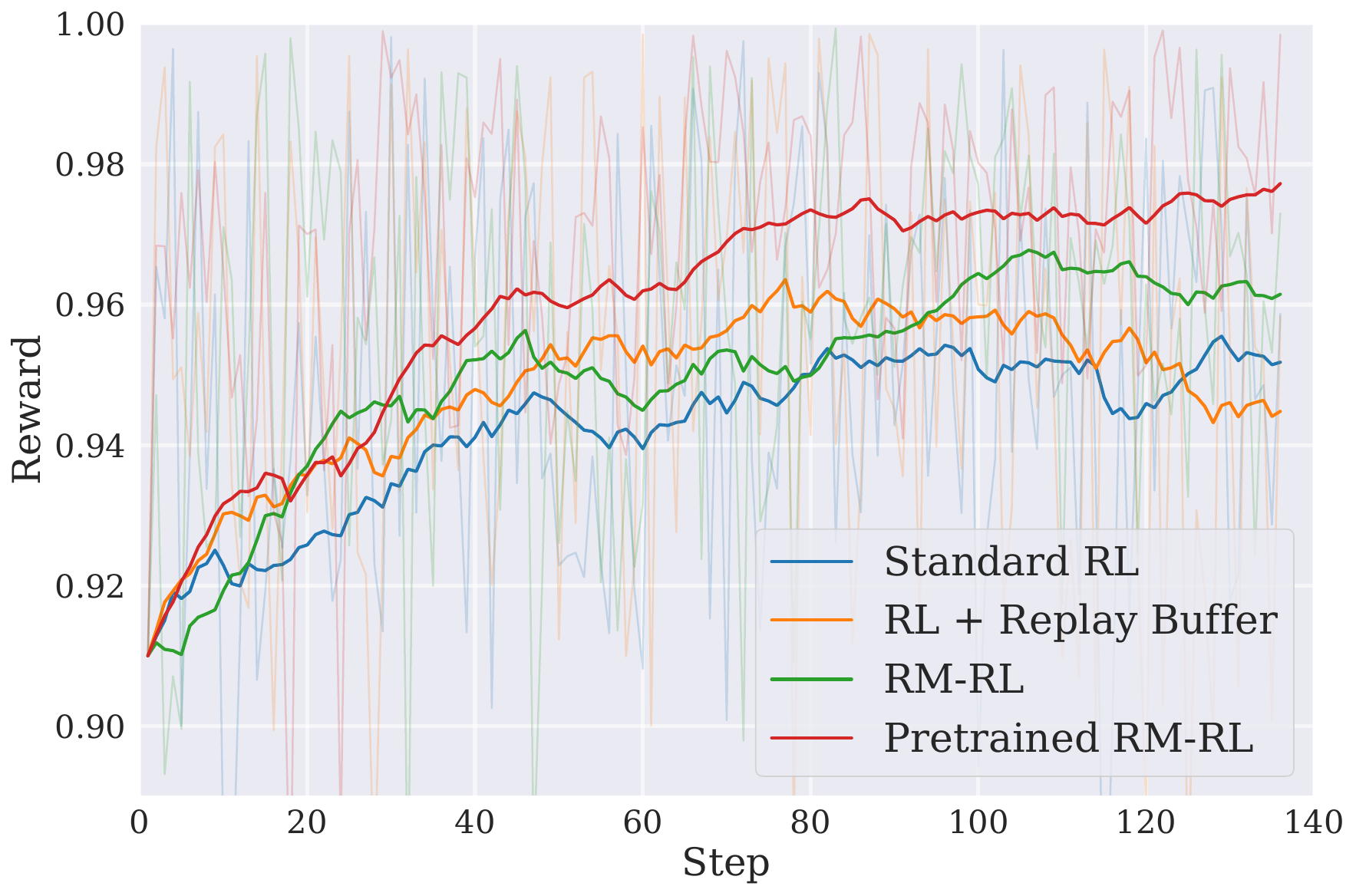}
    \caption{The results of reward curves. The reward curves record the reward changes during real-world RL training from the selected baselines, Standard RL, standard RL with replay buffer (RL + Replay Buffer), the proposed Role-Model Reinforcement Learning (RM-RL), and pretrained RM-RL.}
    \label{fig:exp:rewards_trend}
\end{figure}

We record the reward changes of the selected methods during the RL training, and apply the Exponential Moving Average algorithm to smooth the reward sequences, providing a clearer visualization of the performance trend. Since each trial takes about 2-3 minutes to finish, one successful training may take 7 hours. The results are shown in~\fref{fig:exp:rewards_trend}. As we can see from the picture, the rewards of RM-RL and Pretrained RM-RL increase quickly at the beginning and finally converge to higher reward values than the Standard RL and RL + Replay Buffer methods, demonstrating the effectiveness of the proposed role-model labeling mechanism in improving training efficiency. 
Compared with the reward curves of other RL methods without pretraining, the reward curve of the pretrained RM-RL shows a more consistent and stable improvement, demonstrating that reusing the previous role-model-labeled training data as a pretraining stage is effective to improve the training stability and efficiency. 

\begin{table}[t]
    \centering
    \begin{adjustbox}{width=\columnwidth}
    \begin{tabular}{c|c|c}
    \toprule
    Methods& $T_\tau^{10}$ & Average Reward\\
    \midrule
    Standard RL~\cite{williams1992simple} & 45 & 0.948 (0.029)\\
    RL + Replay Buffer~\cite{mnih2015human} & 36 & 0.953 (0.031)\\
    RM-RL     & 35 & 0.956 (0.027)\\
    Pretrained RM-RL & \textbf{30} & \textbf{0.969} (0.023)\\
    \bottomrule
    \end{tabular}
    \end{adjustbox}
    \caption{Quantitative experimental results of different RL methods evaluated by $T_\tau^{10}$ and average reward.}
    \label{tab:exp:rewards_quantitative_results}
\end{table}
For quantitative analysis, we use two metrics to evaluate the training performance. Similar to the metric $J^{eff.}$~\cite{dulac2019challenges},  we record the training step index at which the learned policy achieves a predefined reward threshold for the tenth occurrence. The calculation of the metric is shown as follows:
\begin{equation}
T_\tau^{10} = \min \left\{ t \in \mathbb{N} \,\middle|\, \sum_{i=1}^t \mathbf{1}_{\{r_i > \tau\}} = 10 \right\},
\end{equation}
where $\tau$ is a predefined reward threshold, $r_i$ is the reward in step $i$, and $\mathbf{1}_{\{r_i > \tau\}}$ is the indicator function, equal to $1$ if $r_i > \tau$ and $0$ otherwise. This metric reflects not only whether the policy can eventually reach the desired reward level, but also how rapidly and consistently it does so across training. By requiring the threshold to be reached multiple times, the metric mitigates the influence of outlier fluctuations and provides a more robust measure of data efficiency and training stability. The second evaluation metric is the average reward, defined as the mean of the rewards obtained over the training steps. This metric reflects the overall training performance of the selected training methods.

\tref{tab:exp:rewards_quantitative_results} summarizes the quantitative experimental results. In particular, Pretrained RM-RL achieves the best results, requiring only 30 steps to reach the threshold for the 10th time (a 33\% reduction from Standard RL) and attaining the highest average reward of 0.969 with the lowest variance, indicating faster convergence and more stable learning.
Although the RL with Relay Buffer demonstrates competitive performance, it exhibits the largest standard deviation in experimental results, indicating instability in training.
\subsection{Real-World Performance}
To evaluate the pick and place performance in the real-world environment, we conduct two experiments. The first experiment follows the training process, putting one cell plate into the given box of a similar size.
We provide 10 random initial positions of the cell plate. The robot needs to pick the cell plate from the initial position and place it in the target position. 
We record the average reward, translation error $e_{trans}$, and rotation error $e_{rot}$ in \tref{tab:exp:real_world_exp1} of the compared methods for evaluation. 
From the experimental results, the Pretrained RM-RL achieves the highest average reward (0.978). Its translational error is reduced to 2.1 mm, representing a 42\% improvement over Standard RL (3.6 mm) and a 64\% reduction compared with RL + Replay Buffer (5.8 mm). For rotational accuracy, Pretrained RM-RL (0.60°) is nearly identical to Standard RL (0.58°) and 41\% better than RL + Replay (1.01°). Overall, these results demonstrate that pretraining using role-model-labeled data significantly enhances real-world pick-and-place performance.

\begin{table}[t]
    \centering
    \begin{adjustbox}{width=\columnwidth}
    \begin{tabular}{c|c|c|c}
    \toprule
    Methods& Average Reward&$e_{trans}$/mm&$e_{rot}$/$\degree$ \\ 
    \midrule
    Standard RL & 0.964 (0.024) & 3.6 (0.27) & \textbf{0.58} (0.49)\\
    RL + Replay Buffer& 0.944 (0.030) & 5.8 (0.33) & 1.01 (0.43) \\
    RM-RL& 0.970 (0.025)& 3.0 (0.30)& 0.89 (0.38)\\
    Pretrained RM-RL& \textbf{0.978} (0.012) & \textbf{2.1} (0.12)& 0.60 (0.45)\\
    \bottomrule
    \end{tabular}
    \end{adjustbox}
    \caption{Performance comparison of different reinforcement learning strategies in real-world robotic picking and placing tasks, evaluated by average reward, position error, and rotation error.}
    \label{tab:exp:real_world_exp1}
\end{table}

The other experiment is placing a cell plate onto the cell plate shelf, given random initial states.
The aim of this experiment is to evaluate whether the predicted pose adjustment is also effective for other tasks that also need precise operations, and the influence of the small errors on these RL-based methods. 
The robotic arm autonomously picks the cell plate with the localization information and the adjustment pose from RL. 
A fixed trajectory is designed to guarantee the cell plate can be put on the cell plate shelf with a proper grasping pose. 
We evaluate the performance of different RL-based methods by measuring the success rate over ten trials.  
The experiment results are summarized in \tref{tab:exp:real_world_exp2}. From the experimental results, the Standard RL achieves a success rate of 50\%, while RL + Replay performs worse with only 40\%, demonstrating the replay with distribution transit error may fail the precise tasks. 
RM-RL also shows unsatisfactory results, achieving only a 70\% success rate, despite presenting results comparable to Pretrained RL + MR in the previous experiment.
The Pretrained RM-RL succeeds in all trials, demonstrating the effectiveness of the proposed receipt with pretraining in enhancing the robot's capability for precise tasks. 

\section{Summary and Future Work}\label{sec:summary and future work}

This paper addresses the challenge of achieving precise robot manipulation in real-world applications, requiring millimeter-level accuracy. 
Online reinforcement learning (RL) can improve safety and accuracy but requires extensive sampling in real-world environments, which is often impractical. 
Offline RL reduces the need for real-world data collection but is hindered by the difficulty of acquiring high-quality datasets and the distribution shift between offline data and real-world execution. 
To overcome these limitations, we proposed Role-Model Reinforcement Learning (RM-RL), a framework that integrates online and offline training. The role-model strategy labels online samples with approximately optimal actions, creating reliable datasets for offline training and policy pretraining. This approach enhances both data and training efficiency. Experimental results demonstrate that RM-RL achieves faster convergence, more stable training, and notable accuracy improvements (53\% in translation and 20\% in rotation) while consistently accomplishing a challenging task: placing a cell plate onto a shelf, which competing RL algorithms fail to perform reliably.

\begin{table}[t]
    \vspace{5pt}
    \centering
    \begin{adjustbox}{width=\columnwidth}

    \begin{tabular}{c|c}
    \toprule
    Methods& Successful Rate \\
    \midrule
    Standard RL  & 50\% \\
    RL + Replay Buffer& 40\% \\
    RM-RL (Ours)& 70\% \\
    Pretrained RM-RL (Ours)& \textbf{100\%} \\
    \bottomrule
    \end{tabular}
    \end{adjustbox}
    \caption{Success rates of different reinforcement learning strategies in real-world picking a cell plate on a shelf.}
    \label{tab:exp:real_world_exp2}
\end{table}

In future work, we will extend RM-RL to more complex and long-horizon robotic tasks that demand sequential decision-making, generalization capability, and sustained precision across multiple stages. Such scenarios will provide a rigorous benchmark to evaluate the scalability of our framework and further establish its potential as a robust, data-efficient solution for real-world robotic applications.

\section{Acknowledgments}
This work is jointly supported by MOE Singapore Tier 1 Grant RG83/25, RS36/24 and a Start-up Grant from Nanyang Technological University.

\balance
\bibliographystyle{assets/plainnat}
\bibliography{references}

\begin{thebibliography}{53}
\providecommand{\natexlab}[1]{#1}
\providecommand{\url}[1]{\texttt{#1}}
\expandafter\ifx\csname urlstyle\endcsname\relax
  \providecommand{\doi}[1]{doi: #1}\else
  \providecommand{\doi}{doi: \begingroup \urlstyle{rm}\Url}\fi

\bibitem[Abdi and Williams(2010)]{abdi2010principal}
Herv{\'e} Abdi and Lynne~J Williams.
\newblock Principal component analysis.
\newblock \emph{Wiley interdisciplinary reviews: computational statistics}, 2\penalty0 (4):\penalty0 433--459, 2010.

\bibitem[Black et~al.(2024)Black, Brown, Driess, Esmail, Equi, Finn, Fusai, Groom, Hausman, Ichter, et~al.]{black2024pi_0}
Kevin Black, Noah Brown, Danny Driess, Adnan Esmail, Michael Equi, Chelsea Finn, Niccolo Fusai, Lachy Groom, Karol Hausman, Brian Ichter, et~al.
\newblock $\pi_0$ : A vision-language-action flow model for general robot control.
\newblock \emph{arXiv preprint arXiv:2410.24164}, 2024.

\bibitem[Busoniu et~al.(2008)Busoniu, Babuska, and De~Schutter]{busoniu2008comprehensive}
Lucian Busoniu, Robert Babuska, and Bart De~Schutter.
\newblock A comprehensive survey of multiagent reinforcement learning.
\newblock \emph{IEEE Transactions on Systems, Man, and Cybernetics, Part C (Applications and Reviews)}, 38\penalty0 (2):\penalty0 156--172, 2008.

\bibitem[Degris et~al.(2012)Degris, Pilarski, and Sutton]{degris2012model}
Thomas Degris, Patrick~M Pilarski, and Richard~S Sutton.
\newblock Model-free reinforcement learning with continuous action in practice.
\newblock In \emph{2012 American control conference (ACC)}, pages 2177--2182. IEEE, 2012.

\bibitem[Dulac-Arnold et~al.(2019)Dulac-Arnold, Mankowitz, and Hester]{dulac2019challenges}
Gabriel Dulac-Arnold, Daniel Mankowitz, and Todd Hester.
\newblock Challenges of real-world reinforcement learning.
\newblock \emph{arXiv preprint arXiv:1904.12901}, 2019.

\bibitem[Fu et~al.(2020)Fu, Kumar, Nachum, Tucker, and Levine]{fu2020d4rl}
Justin Fu, Aviral Kumar, Ofir Nachum, George Tucker, and Sergey Levine.
\newblock D4rl: Datasets for deep data-driven reinforcement learning.
\newblock \emph{arXiv preprint arXiv:2004.07219}, 2020.

\bibitem[Fujimoto et~al.(2019)Fujimoto, Meger, and Precup]{fujimoto2019off}
Scott Fujimoto, David Meger, and Doina Precup.
\newblock Off-policy deep reinforcement learning without exploration.
\newblock In \emph{International conference on machine learning}, pages 2052--2062. PMLR, 2019.

\bibitem[Geng et~al.(2025)Geng, Wang, Wei, Li, Wang, An, Cheng, Lou, Li, Wang, et~al.]{geng2025roboverse}
Haoran Geng, Feishi Wang, Songlin Wei, Yuyang Li, Bangjun Wang, Boshi An, Charlie~Tianyue Cheng, Haozhe Lou, Peihao Li, Yen-Jen Wang, et~al.
\newblock Roboverse: Towards a unified platform, dataset and benchmark for scalable and generalizable robot learning.
\newblock \emph{arXiv preprint arXiv:2504.18904}, 2025.

\bibitem[Geng et~al.(2023)Geng, An, Geng, Chen, Yang, and Dong]{geng2023rlafford}
Yiran Geng, Boshi An, Haoran Geng, Yuanpei Chen, Yaodong Yang, and Hao Dong.
\newblock Rlafford: End-to-end affordance learning for robotic manipulation.
\newblock In \emph{2023 IEEE International conference on robotics and automation (ICRA)}, pages 5880--5886. IEEE, 2023.

\bibitem[Grondman et~al.(2012)Grondman, Busoniu, Lopes, and Babuska]{grondman2012survey}
Ivo Grondman, Lucian Busoniu, Gabriel~AD Lopes, and Robert Babuska.
\newblock A survey of actor-critic reinforcement learning: Standard and natural policy gradients.
\newblock \emph{IEEE Transactions on Systems, Man, and Cybernetics, part C (applications and reviews)}, 42\penalty0 (6):\penalty0 1291--1307, 2012.

\bibitem[Haarnoja et~al.(2018)Haarnoja, Zhou, Hartikainen, Tucker, Ha, Tan, Kumar, Zhu, Gupta, Abbeel, et~al.]{haarnoja2018soft}
Tuomas Haarnoja, Aurick Zhou, Kristian Hartikainen, George Tucker, Sehoon Ha, Jie Tan, Vikash Kumar, Henry Zhu, Abhishek Gupta, Pieter Abbeel, et~al.
\newblock Soft actor-critic algorithms and applications.
\newblock \emph{arXiv preprint arXiv:1812.05905}, 2018.

\bibitem[Hafner et~al.(2019)Hafner, Lillicrap, Ba, and Norouzi]{hafner2019dream}
Danijar Hafner, Timothy Lillicrap, Jimmy Ba, and Mohammad Norouzi.
\newblock Dream to control: Learning behaviors by latent imagination.
\newblock \emph{arXiv preprint arXiv:1912.01603}, 2019.

\bibitem[He et~al.(2016)He, Zhang, Ren, and Sun]{he2016deep}
Kaiming He, Xiangyu Zhang, Shaoqing Ren, and Jian Sun.
\newblock Deep residual learning for image recognition.
\newblock In \emph{Proceedings of the IEEE conference on computer vision and pattern recognition}, pages 770--778, 2016.

\bibitem[Hussein et~al.(2017)Hussein, Gaber, Elyan, and Jayne]{hussein2017imitation}
Ahmed Hussein, Mohamed~Medhat Gaber, Eyad Elyan, and Chrisina Jayne.
\newblock Imitation learning: A survey of learning methods.
\newblock \emph{ACM Computing Surveys (CSUR)}, 50\penalty0 (2):\penalty0 1--35, 2017.

\bibitem[Hwangbo et~al.(2019)Hwangbo, Lee, Dosovitskiy, Bellicoso, Tsounis, Koltun, and Hutter]{hwangbo2019learning}
Jemin Hwangbo, Joonho Lee, Alexey Dosovitskiy, Dario Bellicoso, Vassilios Tsounis, Vladlen Koltun, and Marco Hutter.
\newblock Learning agile and dynamic motor skills for legged robots.
\newblock \emph{Science Robotics}, 4\penalty0 (26):\penalty0 eaau5872, 2019.

\bibitem[James et~al.(2020)James, Ma, Arrojo, and Davison]{james2020rlbench}
Stephen James, Zicong Ma, David~Rovick Arrojo, and Andrew~J Davison.
\newblock Rlbench: The robot learning benchmark \& learning environment.
\newblock \emph{IEEE Robotics and Automation Letters}, 5\penalty0 (2):\penalty0 3019--3026, 2020.

\bibitem[Johannink et~al.(2019)Johannink, Bahl, Nair, Luo, Kumar, Loskyll, Ojea, Solowjow, and Levine]{johannink2019residual}
Tobias Johannink, Shikhar Bahl, Ashvin Nair, Jianlan Luo, Avinash Kumar, Matthias Loskyll, Juan~Aparicio Ojea, Eugen Solowjow, and Sergey Levine.
\newblock Residual reinforcement learning for robot control.
\newblock In \emph{2019 international conference on robotics and automation (ICRA)}, pages 6023--6029. IEEE, 2019.

\bibitem[Kim et~al.(2024)Kim, Pertsch, Karamcheti, Xiao, Balakrishna, Nair, Rafailov, Foster, Lam, Sanketi, et~al.]{kim2024openvla}
Moo~Jin Kim, Karl Pertsch, Siddharth Karamcheti, Ted Xiao, Ashwin Balakrishna, Suraj Nair, Rafael Rafailov, Ethan Foster, Grace Lam, Pannag Sanketi, et~al.
\newblock Openvla: An open-source vision-language-action model.
\newblock \emph{arXiv preprint arXiv:2406.09246}, 2024.

\bibitem[Kirillov et~al.(2023)Kirillov, Mintun, Ravi, Mao, Rolland, Gustafson, Xiao, Whitehead, Berg, Lo, Doll{\'a}r, and Girshick]{kirillov2023segany}
Alexander Kirillov, Eric Mintun, Nikhila Ravi, Hanzi Mao, Chloe Rolland, Laura Gustafson, Tete Xiao, Spencer Whitehead, Alexander~C. Berg, Wan-Yen Lo, Piotr Doll{\'a}r, and Ross Girshick.
\newblock Segment anything.
\newblock \emph{arXiv:2304.02643}, 2023.

\bibitem[Kostrikov et~al.(2021)Kostrikov, Nair, and Levine]{kostrikov2021offline}
Ilya Kostrikov, Ashvin Nair, and Sergey Levine.
\newblock Offline reinforcement learning with implicit q-learning.
\newblock \emph{arXiv preprint arXiv:2110.06169}, 2021.

\bibitem[Kumar et~al.(2020)Kumar, Zhou, Tucker, and Levine]{kumar2020conservative}
Aviral Kumar, Aurick Zhou, George Tucker, and Sergey Levine.
\newblock Conservative q-learning for offline reinforcement learning.
\newblock \emph{Advances in neural information processing systems}, 33:\penalty0 1179--1191, 2020.

\bibitem[Kumar et~al.(2022)Kumar, Hong, Singh, and Levine]{kumar2022should}
Aviral Kumar, Joey Hong, Anikait Singh, and Sergey Levine.
\newblock Should i run offline reinforcement learning or behavioral cloning?
\newblock In \emph{International Conference on Learning Representations}, 2022.

\bibitem[Lan et~al.(2025)Lan, Jiang, Wang, Xie, Zhang, Zhu, Li, Yang, Chen, Gao, et~al.]{lan2025autobio}
Zhiqian Lan, Yuxuan Jiang, Ruiqi Wang, Xuanbing Xie, Rongkui Zhang, Yicheng Zhu, Peihang Li, Tianshuo Yang, Tianxing Chen, Haoyu Gao, et~al.
\newblock Autobio: A simulation and benchmark for robotic automation in digital biology laboratory.
\newblock \emph{arXiv preprint arXiv:2505.14030}, 2025.

\bibitem[Levine et~al.(2020)Levine, Kumar, Tucker, and Fu]{levine2020offline}
Sergey Levine, Aviral Kumar, George Tucker, and Justin Fu.
\newblock Offline reinforcement learning: Tutorial, review, and perspectives on open problems.
\newblock \emph{arXiv preprint arXiv:2005.01643}, 2020.

\bibitem[Li et~al.(2024)Li, Huang, Guo, Wu, Zhang, Zhang, and Ding]{li2024chemistry3d}
Shoujie Li, Yan Huang, Changqing Guo, Tong Wu, Jiawei Zhang, Linrui Zhang, and Wenbo Ding.
\newblock Chemistry3d: Robotic interaction benchmark for chemistry experiments.
\newblock \emph{arXiv preprint arXiv:2406.08160}, 2024.

\bibitem[Lillicrap et~al.(2015)Lillicrap, Hunt, Pritzel, Heess, Erez, Tassa, Silver, and Wierstra]{lillicrap2015continuous}
Timothy~P Lillicrap, Jonathan~J Hunt, Alexander Pritzel, Nicolas Heess, Tom Erez, Yuval Tassa, David Silver, and Daan Wierstra.
\newblock Continuous control with deep reinforcement learning.
\newblock \emph{arXiv preprint arXiv:1509.02971}, 2015.

\bibitem[Liu et~al.(2020{\natexlab{a}})Liu, Dugas, Cesari, Siegwart, and Dub{\'e}]{liu2020robot}
Lucia Liu, Daniel Dugas, Gianluca Cesari, Roland Siegwart, and Renaud Dub{\'e}.
\newblock Robot navigation in crowded environments using deep reinforcement learning.
\newblock In \emph{2020 IEEE/RSJ International Conference on Intelligent Robots and Systems (IROS)}, pages 5671--5677. IEEE, 2020{\natexlab{a}}.

\bibitem[Liu et~al.(2024)Liu, Zeng, Ren, Li, Zhang, Yang, Jiang, Li, Yang, Su, et~al.]{liu2024grounding}
Shilong Liu, Zhaoyang Zeng, Tianhe Ren, Feng Li, Hao Zhang, Jie Yang, Qing Jiang, Chunyuan Li, Jianwei Yang, Hang Su, et~al.
\newblock Grounding dino: Marrying dino with grounded pre-training for open-set object detection.
\newblock In \emph{European conference on computer vision}, pages 38--55. Springer, 2024.

\bibitem[Liu et~al.(2020{\natexlab{b}})Liu, Swaminathan, Agarwal, and Brunskill]{liu2020provably}
Yao Liu, Adith Swaminathan, Alekh Agarwal, and Emma Brunskill.
\newblock Provably good batch off-policy reinforcement learning without great exploration.
\newblock \emph{Advances in neural information processing systems}, 33:\penalty0 1264--1274, 2020{\natexlab{b}}.

\bibitem[Mittal et~al.(2023)Mittal, Yu, Yu, Liu, Rudin, Hoeller, Yuan, Singh, Guo, Mazhar, Mandlekar, Babich, State, Hutter, and Garg]{mittal2023orbit}
Mayank Mittal, Calvin Yu, Qinxi Yu, Jingzhou Liu, Nikita Rudin, David Hoeller, Jia~Lin Yuan, Ritvik Singh, Yunrong Guo, Hammad Mazhar, Ajay Mandlekar, Buck Babich, Gavriel State, Marco Hutter, and Animesh Garg.
\newblock Orbit: A unified simulation framework for interactive robot learning environments.
\newblock \emph{IEEE Robotics and Automation Letters}, 8\penalty0 (6):\penalty0 3740--3747, 2023.
\newblock \doi{10.1109/LRA.2023.3270034}.

\bibitem[Mnih et~al.(2015)Mnih, Kavukcuoglu, Silver, Rusu, Veness, Bellemare, Graves, Riedmiller, Fidjeland, Ostrovski, et~al.]{mnih2015human}
Volodymyr Mnih, Koray Kavukcuoglu, David Silver, Andrei~A Rusu, Joel Veness, Marc~G Bellemare, Alex Graves, Martin Riedmiller, Andreas~K Fidjeland, Georg Ostrovski, et~al.
\newblock Human-level control through deep reinforcement learning.
\newblock \emph{nature}, 518\penalty0 (7540):\penalty0 529--533, 2015.

\bibitem[Mnih et~al.(2016)Mnih, Badia, Mirza, Graves, Lillicrap, Harley, Silver, and Kavukcuoglu]{mnih2016asynchronous}
Volodymyr Mnih, Adria~Puigdomenech Badia, Mehdi Mirza, Alex Graves, Timothy Lillicrap, Tim Harley, David Silver, and Koray Kavukcuoglu.
\newblock Asynchronous methods for deep reinforcement learning.
\newblock In \emph{International conference on machine learning}, pages 1928--1937. PmLR, 2016.

\bibitem[Muratore et~al.(2022)Muratore, Ramos, Turk, Yu, Gienger, and Peters]{muratore2022robot}
Fabio Muratore, Fabio Ramos, Greg Turk, Wenhao Yu, Michael Gienger, and Jan Peters.
\newblock Robot learning from randomized simulations: A review.
\newblock \emph{Frontiers in Robotics and AI}, 9:\penalty0 799893, 2022.

\bibitem[Nair et~al.(2020)Nair, Gupta, Dalal, and Levine]{nair2020awac}
Ashvin Nair, Abhishek Gupta, Murtaza Dalal, and Sergey Levine.
\newblock Awac: Accelerating online reinforcement learning with offline datasets.
\newblock \emph{arXiv preprint arXiv:2006.09359}, 2020.

\bibitem[Puterman(2014)]{puterman2014markov}
Martin~L Puterman.
\newblock \emph{Markov decision processes: discrete stochastic dynamic programming}.
\newblock John Wiley \& Sons, 2014.

\bibitem[Radosavovic et~al.(2024)Radosavovic, Xiao, Zhang, Darrell, Malik, and Sreenath]{radosavovic2024real}
Ilija Radosavovic, Tete Xiao, Bike Zhang, Trevor Darrell, Jitendra Malik, and Koushil Sreenath.
\newblock Real-world humanoid locomotion with reinforcement learning.
\newblock \emph{Science Robotics}, 9\penalty0 (89):\penalty0 eadi9579, 2024.

\bibitem[Rafailov et~al.(2024)Rafailov, Hatch, Singh, Smith, Kumar, Kostrikov, Hansen-Estruch, Kolev, Ball, Wu, et~al.]{rafailov2024d5rl}
Rafael Rafailov, Kyle Hatch, Anikait Singh, Laura Smith, Aviral Kumar, Ilya Kostrikov, Philippe Hansen-Estruch, Victor Kolev, Philip Ball, Jiajun Wu, et~al.
\newblock D5rl: Diverse datasets for data-driven deep reinforcement learning.
\newblock \emph{arXiv preprint arXiv:2408.08441}, 2024.

\bibitem[Ravichandar et~al.(2020)Ravichandar, Polydoros, Chernova, and Billard]{ravichandar2020recent}
Harish Ravichandar, Athanasios~S Polydoros, Sonia Chernova, and Aude Billard.
\newblock Recent advances in robot learning from demonstration.
\newblock \emph{Annual review of control, robotics, and autonomous systems}, 3\penalty0 (1):\penalty0 297--330, 2020.

\bibitem[Rowland et~al.(2019)Rowland, Dadashi, Kumar, Munos, Bellemare, and Dabney]{rowland2019statistics}
Mark Rowland, Robert Dadashi, Saurabh Kumar, R{\'e}mi Munos, Marc~G Bellemare, and Will Dabney.
\newblock Statistics and samples in distributional reinforcement learning.
\newblock In \emph{International Conference on Machine Learning}, pages 5528--5536. PMLR, 2019.

\bibitem[Sarker(2021)]{sarker2021machine}
Iqbal~H Sarker.
\newblock Machine learning: Algorithms, real-world applications and research directions.
\newblock \emph{SN computer science}, 2\penalty0 (3):\penalty0 160, 2021.

\bibitem[Schulman et~al.(2017)Schulman, Wolski, Dhariwal, Radford, and Klimov]{schulman2017proximal}
John Schulman, Filip Wolski, Prafulla Dhariwal, Alec Radford, and Oleg Klimov.
\newblock Proximal policy optimization algorithms.
\newblock \emph{arXiv preprint arXiv:1707.06347}, 2017.

\bibitem[Sutton et~al.(1999)Sutton, McAllester, Singh, and Mansour]{sutton1999policy}
Richard~S Sutton, David McAllester, Satinder Singh, and Yishay Mansour.
\newblock Policy gradient methods for reinforcement learning with function approximation.
\newblock \emph{Advances in neural information processing systems}, 12, 1999.

\bibitem[Tobin et~al.(2017)Tobin, Fong, Ray, Schneider, Zaremba, and Abbeel]{tobin2017domain}
Josh Tobin, Rachel Fong, Alex Ray, Jonas Schneider, Wojciech Zaremba, and Pieter Abbeel.
\newblock Domain randomization for transferring deep neural networks from simulation to the real world.
\newblock In \emph{2017 IEEE/RSJ international conference on intelligent robots and systems (IROS)}, pages 23--30. IEEE, 2017.

\bibitem[Todorov et~al.(2012)Todorov, Erez, and Tassa]{todorov2012mujoco}
Emanuel Todorov, Tom Erez, and Yuval Tassa.
\newblock Mujoco: A physics engine for model-based control.
\newblock In \emph{2012 IEEE/RSJ international conference on intelligent robots and systems}, pages 5026--5033. IEEE, 2012.

\bibitem[Walke et~al.(2023)Walke, Black, Zhao, Vuong, Zheng, Hansen-Estruch, He, Myers, Kim, Du, et~al.]{walke2023bridgedata}
Homer~Rich Walke, Kevin Black, Tony~Z Zhao, Quan Vuong, Chongyi Zheng, Philippe Hansen-Estruch, Andre~Wang He, Vivek Myers, Moo~Jin Kim, Max Du, et~al.
\newblock Bridgedata v2: A dataset for robot learning at scale.
\newblock In \emph{Conference on Robot Learning}, pages 1723--1736. PMLR, 2023.

\bibitem[Wan et~al.(2023)Wan, Geng, Liu, Shan, Yang, Yi, and Wang]{wan2023unidexgrasp++}
Weikang Wan, Haoran Geng, Yun Liu, Zikang Shan, Yaodong Yang, Li~Yi, and He~Wang.
\newblock Unidexgrasp++: Improving dexterous grasping policy learning via geometry-aware curriculum and iterative generalist-specialist learning.
\newblock In \emph{Proceedings of the IEEE/CVF International Conference on Computer Vision}, pages 3891--3902, 2023.

\bibitem[Wijayathunga et~al.(2023)Wijayathunga, Rassau, and Chai]{wijayathunga2023challenges}
Liyana Wijayathunga, Alexander Rassau, and Douglas Chai.
\newblock Challenges and solutions for autonomous ground robot scene understanding and navigation in unstructured outdoor environments: A review.
\newblock \emph{Applied Sciences}, 13\penalty0 (17):\penalty0 9877, 2023.

\bibitem[Williams(1992)]{williams1992simple}
Ronald~J Williams.
\newblock Simple statistical gradient-following algorithms for connectionist reinforcement learning.
\newblock \emph{Machine learning}, 8\penalty0 (3):\penalty0 229--256, 1992.

\bibitem[Xu et~al.(2023)Xu, Wan, Zhang, Liu, Shan, Shen, Wang, Geng, Weng, Chen, et~al.]{xu2023unidexgrasp}
Yinzhen Xu, Weikang Wan, Jialiang Zhang, Haoran Liu, Zikang Shan, Hao Shen, Ruicheng Wang, Haoran Geng, Yijia Weng, Jiayi Chen, et~al.
\newblock Unidexgrasp: Universal robotic dexterous grasping via learning diverse proposal generation and goal-conditioned policy.
\newblock In \emph{Proceedings of the IEEE/CVF Conference on Computer Vision and Pattern Recognition}, pages 4737--4746, 2023.

\bibitem[Zhang(2002)]{zhang2002flexible}
Zhengyou Zhang.
\newblock A flexible new technique for camera calibration.
\newblock \emph{IEEE Transactions on pattern analysis and machine intelligence}, 22\penalty0 (11):\penalty0 1330--1334, 2002.

\bibitem[Zhao et~al.(2020)Zhao, Queralta, and Westerlund]{zhao2020sim}
Wenshuai Zhao, Jorge~Pe{\~n}a Queralta, and Tomi Westerlund.
\newblock Sim-to-real transfer in deep reinforcement learning for robotics: a survey.
\newblock In \emph{2020 IEEE symposium series on computational intelligence (SSCI)}, pages 737--744. IEEE, 2020.

\bibitem[Zhou et~al.(2022)Zhou, Ke, Srinivasa, Gupta, Rajeswaran, and Kumar]{zhou2022real}
Gaoyue Zhou, Liyiming Ke, Siddhartha Srinivasa, Abhinav Gupta, Aravind Rajeswaran, and Vikash Kumar.
\newblock Real world offline reinforcement learning with realistic data source.
\newblock \emph{arXiv preprint arXiv:2210.06479}, 2022.

\bibitem[Zitkovich et~al.(2023)Zitkovich, Yu, Xu, Xu, Xiao, Xia, Wu, Wohlhart, Welker, Wahid, et~al.]{zitkovich2023rt}
Brianna Zitkovich, Tianhe Yu, Sichun Xu, Peng Xu, Ted Xiao, Fei Xia, Jialin Wu, Paul Wohlhart, Stefan Welker, Ayzaan Wahid, et~al.
\newblock Rt-2: Vision-language-action models transfer web knowledge to robotic control.
\newblock In \emph{Conference on Robot Learning}, pages 2165--2183. PMLR, 2023.

\end{thebibliography}
\addtolength{\textheight}{-12cm}  
\end{document}